# Dynamic Loss Balancing for Joint SOH and RUL Prediction of Lithium-Ion Batteries via a Rotary SOH-Injected Prior Battery Transformer

Shuhao Chen[1], Tianyu Shi[1], Yiwen Huang[1], Chengyi Tu[1]

[1]School of Economics and Management, Zhejiang Sci-Tech University, Hangzhou, 310018, China.

Corresponding author. Email: chengyitu1986@gmail.com

## Abstract

The deployment of reliable lithium-ion battery management systems is crucial for accelerating electrification, yet the joint prognosis of State of Health (SOH) and Remaining Useful Life (RUL) remains severely hindered by task heteroscedasticity. Conventional multi-task learning frameworks fail to balance the bounded, low-variance noise of SOH estimation with the unbounded, nonlinearly expanding uncertainty of long-term RUL predictions. Here, we present the Rotary SOH-Injected Prior Battery Transformer (RoSIP-Batt), a unified co-estimation framework that resolves these optimization conflicts. By formulating joint prediction as a Bayesian multi-task objective, RoSIP-Batt introduces a homoscedastic uncertainty weighting mechanism to dynamically scale task-specific gradients based on learned residual noise levels. The architecture leverages decoupled dual classification tokens and a per-dimension gated fusion mechanism, secured by a gradient-detachment operator to prevent high-variance RUL updates from corrupting the stable SOH representation space. To capture electrochemical degradation patterns without relying on absolute cycle steps, Rotary Position Embedding (RoPE) is incorporated into a shared Transformer backbone to model translation-invariant relative temporal profiles. Crucially, the intermediate SOH estimate is directly injected into the RUL regression head as a physical degradation prior. Evaluations across the NASA, MIT-Stanford, and HUST datasets show that RoSIP-Batt significantly outperforms state-of-the-art baselines, reducing SOH estimation error to 1.994% MAE on NASA and restricting RUL prediction error to 62.85 cycles on Stanford. These findings establish RoSIP-Batt as a highly generalizable, computationally efficient solution suitable for real-time embedded BMS deployment.

# 1.Introduction

Lithium-ion batteries have emerged as the cornerstone of modern energy storage systems[1], driving the global transition toward electric vehicles (EVs)[2] and smart grids. However, continuous physical and electrochemical degradation during cycling remains an inherent challenge[3], causing gradual capacity fade and elevating the risk of catastrophic battery failures[4]. In the field of battery health management[5], State of Health (SOH) and Remaining Useful Life (RUL) represent the two most critical metrics of battery degradation[6]. SOH reflects the current maximum capacity relative to the nominal capacity[7], while RUL estimates the number of remaining cycles before the battery reaches its end-of-life (EOL) threshold[8].

Traditionally, data-driven prognostic methodologies have treated SOH and RUL estimation as decoupled[9,10], independent tasks[11,12]. Classical machine learning methods, such as Gaussian Process Regression (GPR)[13,14], and deep learning architectures, such as Convolutional Neural Networks (CNNs)[15] and Long Short-Term Memory (LSTM) networks[16], have demonstrated efficacy when applied to these tasks individually[17]. However, such independent modeling paradigms overlook the fundamental electrochemical reality: both metrics are macroscopic manifestations of the same microscopic degradation processes[18], including the growth of the solid electrolyte interphase (SEI) layer[11], lithium inventory depletion, and active material dissolution[19]. Multi-Task Learning (MTL) addresses this limitation by employing a shared feature extractor to learn joint representations[20], thereby improving generalizability and computational efficiency[11].

Despite its conceptual advantages[21], the application of MTL in battery prognostics is fundamentally constrained by task-specific heteroscedasticity[22]. SOH is a strictly bounded ratio characterized by low-variance, homogeneous measurement noise[23]. Conversely, RUL is a wider-range, unbounded temporal count where prediction uncertainty amplifies nonlinearly as the cell approaches EOL[24]. Conventional MTL paradigms rely on static, heuristic loss-weighting schemes that linearly combine task-specific losses using fixed coefficients. Such static approaches are structurally rigid and fail to adapt to the dynamic noise profiles of individual tasks, typically resulting in severe gradient imbalance and task domination[25]. Consequently, optimization trajectories tend to favor the more easily minimizable SOH loss, culminating in suboptimal RUL convergence[26]. Furthermore, existing deep learning models struggle with cross-dataset generalization, frequently failing when evaluated on testing protocols or chemical systems unseen during training[27,28].

To address these limitations, this paper introduces the Rotary SOH-Injected Prior Battery Transformer (RoSIP-Batt) framework for joint SOH and RUL co-estimation[29]. By formulating the joint prediction objective within a unified Bayesian framework, we incorporate a homoscedastic uncertainty weighting mechanism that dynamically balances task-specific gradients without requiring manual hyperparameter tuning[30]. Concurrently, the core shared Transformer encoder is integrated with Rotary Position Embedding (RoPE)[31] to capture translation-invariant degradation trajectories across lengthy and noisy charging cycles[32].

Specifically, our framework incorporates several key architectural innovations. We design Dual Class (CLS) Tokens, defining separate SOH and RUL CLS tokens to encapsulate distinct task semantics[33]. A Per-Dimension Gated Fusion mechanism is introduced to dynamically combine SOH and RUL representation channels via a learned sigmoid gating operator[34,35]. Additionally, a Gradient Detachment strategy is implemented to decouple the RUL backpropagation pathway from SOH feature extraction[36], thereby shielding the stable SOH representations from high-variance RUL gradients. Finally, we incorporate a Physical SOH Injection mechanism, wherein the intermediate estimated SOH is directly fed into the RUL regression head to serve as a physical degradation constraint prior[26].

The principal contributions of this work are summarized as follows. First, we propose a novel Transformer-based multi-task learning architecture featuring physics-constrained regression heads (Sigmoid and Softplus) specifically tailored to the boundaries of physical battery health metrics[37,38]. Second, we develop a Bayesian homoscedastic uncertainty-weighted loss function that dynamically adapts loss gradients, successfully resolving the optimization conflict arising from SOH-RUL task heteroscedasticity. Third, we conduct extensive validations across NASA, MIT-Stanford, and HUST datasets, demonstrating state-of-the-art co-estimation accuracy, exceptional zero-shot cross-dataset transferability, and high robustness under partial charging scenarios[39].

# 2.Methods

To address the challenge of task heteroscedasticity and ensure electrochemical consistency in joint battery diagnostics, the proposed RoSIP-Batt framework leverages a modular, physics-constrained multi-task learning architecture, as schematically illustrated in Figure 1. The overall pipeline proceeds as follows: multi-channel charging profiles are first processed by a one-dimensional convolutional frontend to extract local temporal features, and then encoded by a shared Transformer equipped with Rotary Position Embedding (RoPE) to capture relative long-term degradation dependencies. Decoupled dual classification tokens and a per-dimension gated fusion layer represent task-specific semantics, while a gradient-detachment operator shields the stable SOH latent space from high-variance RUL gradients. The intermediate SOH estimate is subsequently injected into the RUL regression head as a physical constraint prior, and the joint network is optimized via a Bayesian homoscedastic uncertainty loss.

**Uncertainty-Weighted Multitask Transformer for Joint SOH and RUL Prediction**

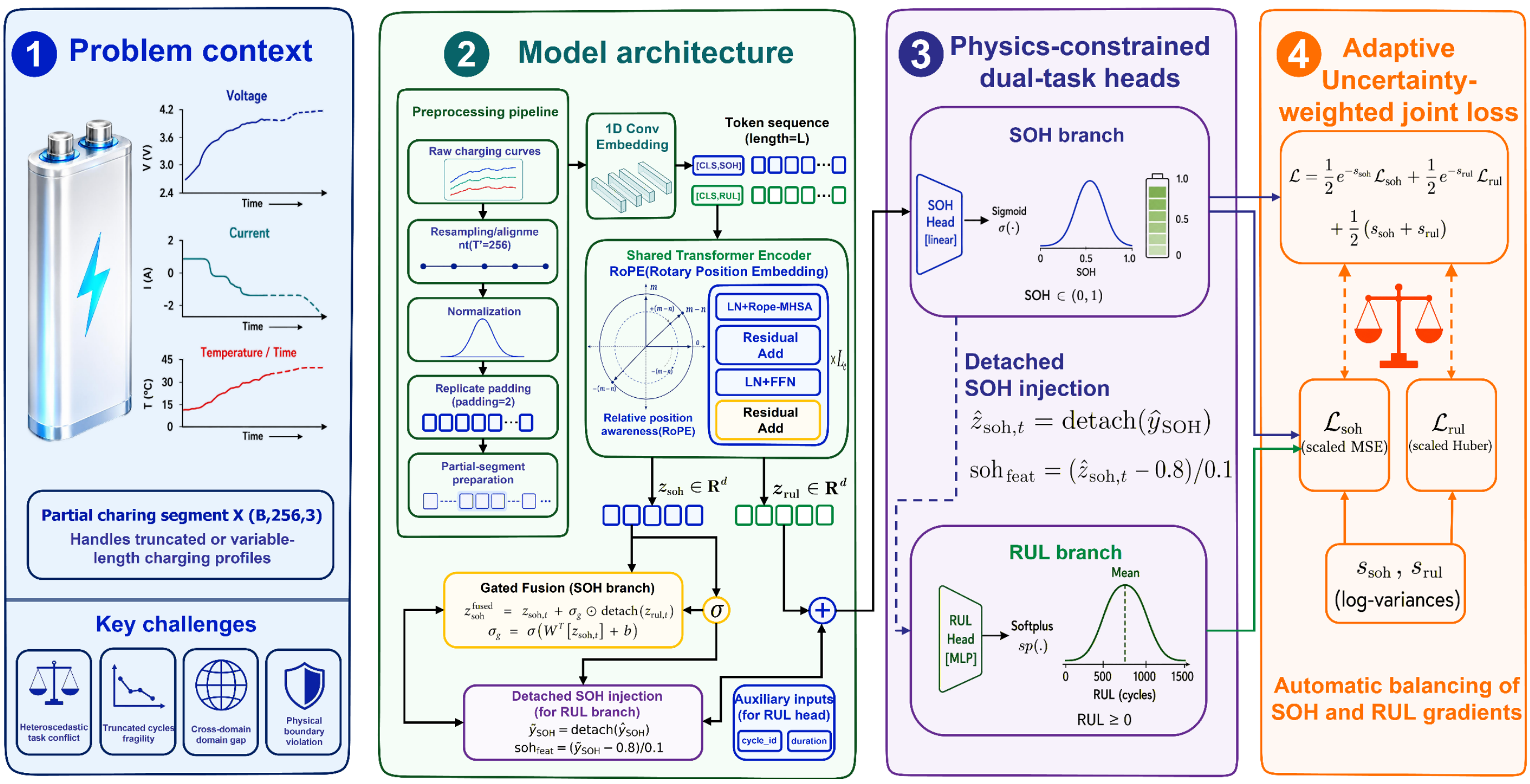

**Figure 1. RoSIP-Batt Model Architecture and Information Flow.** Comprehensive architectural schematic of the proposed Rotary SOH-Injected Prior Battery Transformer (RoSIP-Batt) framework. The raw multi-channel charging time-series (Voltage, Current, Temperature) are first processed by a Conv1D frontend with Batch Normalization and GELU activation to extract local features. The sequence is prepended with Dual CLS tokens ( $\mathbf{cls}_{SOH}$ and $\mathbf{cls}_{RUL}$ ) and encoded using a shared Pre-LN Transformer with Rotary Position Embedding (RoPE) to capture long-term relative temporal dependencies. The SOH and RUL representations are combined using a per-dimension Gated Fusion mechanism with a gradient-detach operator to insulate SOH from noisy RUL updates. SOH predictions are constrained by a Logistic Sigmoid head, while RUL predictions are constrained by a Softplus head with physical SOH and chrono feature injection.

## 2.1 Probabilistic Formulation and Data Alignment

Let the multivariate charging profile sequence for a given battery cycle be denoted by $\mathbf{X} \in \mathbb{R}^{T \times C}$, where $T = 256$ represents the uniform resampled sequence length applied to all charging curves, and $C = 3$ represents the input channels: Voltage ( $V$ ), Current ( $I$ ), and Temperature ( $T$ ). Our objective is to construct a deep neural network $f^{W}(\mathbf{X})$ parameterized by weights $W$ to jointly estimate SOH ( $y_{\mathrm{SOH}}$ ) and RUL ( $y_{\mathrm{RUL}}$ ). To prevent data leakage, all datasets are standardized and train/test splits are strictly enforced at the individual cell level. We formulate the joint prediction task as a Maximum Likelihood Estimation (MLE) problem under a Bayesian perspective, assuming that the target variables are corrupted by task-specific independent homoscedastic Gaussian noise:

$$y_t \sim \mathcal{N}(f_t^{W}(\mathbf{X}), \sigma_t^2) \quad \text{for } t \in \{\mathrm{SOH}, \mathrm{RUL}\} \tag{1}$$

where $\sigma_t^2$ denotes the task-specific observation noise variance.

## 2.2 CNN Frontend and Shared Transformer Encoder Architecture

To extract robust local temporal features and filter high-frequency sensor noise, the input sequence $\mathbf{X}$ is first processed by a one-dimensional convolutional (Conv1D) frontend:

$$\mathbf{H}_0 = \text{GELU}(\text{BatchNorm1d}(\text{Conv1d}(\mathbf{X}^\top)))^\top \tag{2}$$

which maps the input channels from $C=3$ to a $d=128$ dimensional hidden space. The core feature extractor consists of $L=2$ stacked Transformer encoder blocks. To guarantee numerical stability and stable training convergence without requiring a learning rate warmup phase, we employ a Pre-Layer Normalization (Pre-LN) architecture. Each encoder block comprises a Multi-Head Attention (MHA) module configured with $h=4$ heads, followed by a Feed-Forward Network (FFN). The FFN utilizes a $4\times$ expansion ratio, projecting the feature space to a hidden dimension of 512 before returning to $d=128$ via a Gaussian Error Linear Unit (GELU) activation. A dropout rate of 0.1 is applied uniformly across both the MHA and FFN modules to mitigate overfitting on small-scale battery cycling datasets.

## 2.3 Dual CLS Tokens and Per-Dimension Gated Fusion

To preserve individual task semantics and prevent negative transfer, our architecture incorporates Dual Class (CLS) tokens, denoted as $\mathbf{cls}_{\text{SOH}}$ and $\mathbf{cls}_{\text{RUL}} \in \mathbb{R}^{1\times d}$, in place of a single shared token. These tokens are prepended to the sequence embeddings $\mathbf{H}_0$ prior to the Transformer blocks:

$$\mathbf{H} = \left[\mathbf{cls}_{\text{SOH}}; \mathbf{cls}_{\text{RUL}}; \mathbf{H}_0\right] \in \mathbb{R}^{(T+2)\times d} \tag{3}$$

Following propagation through the Transformer blocks and a final LayerNorm layer, the task-specific representations $\mathbf{z}_{\text{SOH}}$ and $\mathbf{z}_{\text{RUL}}$ are extracted from the corresponding output tokens. To facilitate stable interaction between the tasks while mitigating optimization interference, we propose a Gradient Detachment strategy coupled with a Per-Dimension Gated Fusion mechanism. Because SOH features exhibit stable, short-term trends whereas RUL features suffer from high, long-term variance, we detach the gradient flow of the RUL pathway before it is integrated into the SOH path:

$$\mathbf{z}_{\text{RUL}}^{\text{ref}} = \text{detach}(\mathbf{z}_{\text{RUL}}) \tag{4}$$

A per-dimension sigmoid gate is then computed to dynamically regulate the fusion of RUL information:

$$\mathbf{g} = \sigma(\mathbf{W}_g \mathbf{z}_{\text{SOH}} + \mathbf{b}_g) \in \mathbb{R}^d \tag{5}$$

$$\mathbf{z}_{\text{SOH}}^{\text{fused}} = \mathbf{z}_{\text{SOH}} + \mathbf{g} \odot \mathbf{z}_{\text{RUL}}^{\text{ref}} \tag{6}$$

where $\odot$ denotes the Hadamard (element-wise) product, and $\mathbf{W}_g$ and $\mathbf{b}_g$ represent learnable parameters. The gate bias $\mathbf{b}_g$ is initialized to -3.0, which biases the gate to remain closed during early training phases, forcing the network to prioritize independent SOH features before gradually learning the optimal task-fusion ratio.

## 2.4 Rotary Position Embedding (RoPE) in the Complex Domain

In contrast to standard Transformer models that rely on absolute position encodings, our framework integrates Rotary Position Embedding (RoPE) applied exclusively to the Query and Key vectors within the MHA modules. This formulation is highly advantageous for battery health monitoring because degradation features—such as voltage relaxation rates and charging current profiles—are characterized by translation-invariant, relative temporal changes rather than absolute cycle step positions. For time steps $m$ and $n$, and respective query/key vectors $\mathbf{q}_m, \mathbf{k}_n \in \mathbb{R}^d$, RoPE projects these vectors into the complex space. By applying a rotation factor $e^{im\theta}$, the transformed representations are defined as $\tilde{\mathbf{q}}_m = \mathbf{q}_m e^{im\theta}$ and $\tilde{\mathbf{k}}_n = \mathbf{k}_n e^{in\theta}$. The attention inner product is thus derived as:

$$\langle \tilde{\mathbf{q}}_m, \tilde{\mathbf{k}}_n \rangle = \mathrm{Re}\left(\mathbf{q}_m \mathbf{k}_n^* e^{i(m-n)\theta}\right) \tag{7}$$

where $*$ denotes the complex conjugate, and $\mathrm{Re}(\cdot)$ extracts the real component. This formulation guarantees that the attention mechanism is exclusively parameterized by the relative distance $(m-n)$. In practice, this is implemented via a block-diagonal rotation matrix $R_{\Theta,m}$ applied across $d/2$ dimension pairs, with the base frequency set to $\theta_j = 10000^{-2j/d}$.

## 2.5 Physics-Constrained Dual-Task Heads with SOH Injection

For the SOH regression head, since State of Health represents a bounded capacity ratio ($y_{\mathrm{SOH}} \in (0,1]$), we enforce physical range constraints using a Logistic Sigmoid activation function:

$$\hat{y}_{\mathrm{SOH}} = \mathrm{Sigmoid}(\mathbf{W}_S \mathbf{z}_{\mathrm{SOH}}^{\mathrm{fused}} + b_S) \tag{8}$$

For the RUL head, along with the RUL latent representation, we incorporate key temporal indicators (Chrono features representing absolute cycle index and charging duration) and the estimated SOH to guide long-term cycle predictions:

$$\mathbf{z}_{\mathrm{RUL}}^{\mathrm{concat}} = \left[\mathbf{z}_{\mathrm{RUL}}; \mathrm{cycle}_{\mathrm{idx}}; \mathrm{duration}; \mathrm{SOH}_{\mathrm{feat}}\right] \tag{9}$$

where the SOH feature is scaled and centered as $\mathrm{SOH}_{\mathrm{feat}} = (\hat{y}_{\mathrm{SOH}} - 0.8)/0.1$. The RUL prediction is then computed using a Softplus activation to ensure non-negativity:

$$\hat{y}_{\mathrm{RUL}} = \mathrm{Softplus}(\mathbf{W}_R \mathbf{z}_{\mathrm{RUL}}^{\mathrm{concat}} + b_R) \tag{10}$$

where $\mathbf{W}_S, b_S, \mathbf{W}_R$, and $b_R$ represent task-specific weights and biases. This direct injection of SOH predictions into the RUL regression pathway enforces physical consistency between current capacity degradation and the remaining operational life, motivating the "SOH-Injected Prior" designation of the proposed RoSIP-Batt framework.

## 2.6 Homoscedastic Uncertainty-Weighted Joint Loss

To address gradient imbalances caused by task heteroscedasticity, we formulate the optimization problem as the minimization of the negative log-likelihood (NLL) of a joint Gaussian probability distribution. Given the bounded nature of SOH and the unbounded, wide-range variance of RUL, their task-specific loss functions are defined distinctively: SOH loss ( $\mathcal{L}_{\text{SOH}}$ ) is computed via the Mean Squared Error (MSE), whereas RUL loss ( $\mathcal{L}_{\text{RUL}}$ ) utilizes the Huber loss parameterized by a threshold $\delta$ to mitigate outlier sensitivity:

$$\mathcal{L}_{\text{RUL}} = \text{Huber}(\hat{y}_{\text{RUL}}, y_{\text{RUL}}, \delta) \tag{11}$$

Assuming conditional task independence given the model parameters $W$, the joint negative log-likelihood is formulated as:

$$-\log p(y_{\text{SOH}}, y_{\text{RUL}} \mid \mathbf{X}, W, \sigma_{\text{SOH}}, \sigma_{\text{RUL}}) \propto \frac{1}{2\sigma_{\text{SOH}}^2}\mathcal{L}_{\text{SOH}} + \frac{1}{2\sigma_{\text{RUL}}^2}\mathcal{L}_{\text{RUL}} + \log(\sigma_{\text{SOH}}) + \log(\sigma_{\text{RUL}}) \tag{12}$$

where $\sigma_{\text{SOH}}$ and $\sigma_{\text{RUL}}$ represent the task-specific residual noise scales. To avoid numerical instability during the optimization of $\sigma_t^2$, we define learnable log-variance parameters, initialized to $0.0$, as $s_t = \log(\sigma_t^2)$. Substituting these parameters into the joint NLL yields our final dynamic loss function:

$$\mathcal{L}(W, s_{\text{SOH}}, s_{\text{RUL}}) = \frac{1}{2}\exp(-s_{\text{SOH}})\mathcal{L}_{\text{SOH}} + \frac{1}{2}\exp(-s_{\text{RUL}})\mathcal{L}_{\text{RUL}} + \frac{1}{2}s_{\text{SOH}} + \frac{1}{2}s_{\text{RUL}} \tag{13}$$

This probabilistic formulation allows the optimizer to dynamically adjust the loss weights during training. When RUL predictions exhibit high variance, the network autonomously attenuates the corresponding penalty scale $\exp(-s_{\text{RUL}})$, protecting the shared feature representations of the stable SOH task from gradient disruption.

# 3. Experimental Setup and Results

To evaluate the predictive accuracy, cross-chemistry generalizability, and computational efficiency of RoSIP-Batt, we conduct systematic experiments across three diverse public battery datasets representing distinct chemical compositions, cell structures, and cycling protocols. The following subsections describe the dataset specifications, baseline architectures, implementation details, quantitative benchmark comparisons, ablation analyses, and Huber threshold sensitivity studies.

## 3.1 Datasets and Technical Parameters

To evaluate the predictive accuracy and generalizability of RoSIP-Batt, comprehensive experiments were conducted across three public datasets representing distinct battery chemistries, operational conditions, and cycling lifespans. First, the NASA PCoE Dataset comprises cylindrical NMC/Graphite cells cycled under a standard Constant Current-Constant Voltage (CC-CV) charging protocol (1.5 A CC up to 4.2 V, with a CV cut-off current of 20 mA) and a constant-current discharge profile. Second, the MIT-Stanford Dataset contains LFP/Graphite pouch cells subjected to extreme fast-charging profiles (ranging from 4C to 6C charging rates to 3.6 V), resulting in wide variations in end-of-life (EOL) cycle counts (spanning 150 to over 2,300 cycles). Third, the HUST Dataset features LFP/Graphite cells cycled under highly dynamic multi-stage constant current charging profiles, designed to replicate practical charging variability. Detailed configurations for each dataset are summarized in Table 1.

**Table 1: Technical specifications and cycling protocols of the SOH and RUL degradation datasets.**

| Dataset | Chemistry | Capacity (Ah) | Charging Protocol | Discharging Protocol | Validation Method | Scale |
|---|---|---|---|---|---|---|
| NASA PCoE | NMC/Graphite | 2 | CC-CV (1.5 A CC, CV to 4.2 V) | CC (2.0 A CC, cut-off 2.7 V) | LOOCV | 4 cells |
| Stanford | LFP/Graphite | 1.1 | Fast CC (e.g., 4C to 3.6C) | CC (1.1 A CC, cut-off 2.0 V) | 10-seed split | 41 cells |
| HUST | LFP/Graphite | 1.1 | Multi-stage CC | CC (1.1 A CC, cut-off 2.5 V) | 5-seed split | 5 seeds |

## 3.2 Baseline Architectures

We benchmarked our model against two categories of baseline architectures spanning single-task sequence models and conventional multi-task learning frameworks. The first category comprises two single-task models that predict SOH and RUL independently. The Vanilla Transformer adopts the standard encoder-only Transformer architecture with learnable absolute position embeddings and a single [CLS] token, trained separately for each target metric. The 1D-CNN employs stacked one-dimensional convolutional layers with max-pooling and fully connected heads, serving as a representative local feature extractor that captures short-range temporal patterns without explicit long-range attention. The second category comprises three multi-task learning frameworks that jointly predict both metrics through shared representations. The Multi-gate Mixture-of-Experts (MMoE) model replaces the conventional shared backbone with multiple expert sub-networks, each gated by task-specific softmax routing functions that learn to selectively combine expert outputs for SOH and RUL prediction. The Progressive Layered Extraction (PLE) model extends the mixture-of-experts paradigm by introducing task-specific expert modules alongside shared experts, organized in a progressive extraction architecture that explicitly separates shared and task-private knowledge across multiple fusion layers to mitigate the seesaw effect between competing objectives. The Multi-Task Attention Network (MTAN) augments a shared encoder with task-specific soft-attention modules that dynamically attend to different feature subspaces for each task, enabling fine-grained, attention-driven task specialization. Together, these five baselines provide comprehensive coverage of shared-parameter, gated mixture-of-experts, progressive extraction, and attention-based MTL paradigms, enabling a rigorous and fair evaluation of the proposed RoSIP-Batt framework.

## 3.3 Implementation Details and Computational Environment

To guarantee strict reproducibility, all experiments were conducted on a high-performance computing node equipped with dual Intel Xeon Platinum 8469C processors ($2 \times 48$ physical cores, $3.1\,\text{GHz}$ base / $3.8\,\text{GHz}$ turbo, 192 logical threads) and 8 NVIDIA H20 Tensor Core GPUs. The Distributed Data Parallel (DDP) framework was employed to distribute the training workload uniformly across all 8 GPUs, accelerating convergence on the pooled multi-source dataset. CPU inference latency was benchmarked on a single Xeon core under sequential execution to approximate the computational conditions of resource-constrained edge processors. The proposed RoSIP-Batt and all benchmark architectures were implemented in Python 3.11 with PyTorch (v2.0+) and CUDA acceleration. The network parameters were optimized utilizing the AdamW optimizer with an initial learning rate of $1\times10^{-3}$, a weight decay of 0.01, and a cosine annealing learning rate scheduler. Training was conducted for 300 epochs with a batch size of 32. Data preprocessing and statistical evaluations were performed using NumPy and scikit-learn. To prevent temporal data leakage, train/test splits were strictly enforced at the individual cell level.

## 3.4 Benchmark Performance Comparison

The performance metrics across the NASA, Stanford, and HUST datasets are summarized in Table 2. Our proposed RoSIP-Batt model consistently achieves state-of-the-art results.

**Table 2: Quantitative benchmark comparison of SOH estimation error (MAE, %) and RUL prediction error (MAE, cycles) across NASA, MIT-Stanford, and HUST datasets.**

| Dataset | Model Family | Model | SOH MAE (%) | RUL MAE (cycles) |
|---|---|---|---|---|
| NASA PCoE (NMC, LOOCV) | Single-task time-series baseline | Vanilla Transformer | 3.053 ± 1.396 | 28.19 ± 14.11 |
| | | CNN-1D | 3.479 ± 1.621 | 26.21 ± 3.19 |
| | Multi-task learning baseline | MMoE | 2.823 ± 0.375 | 29.29 ± 2.30 |
| | | PLE | 2.909 ± 0.486 | 22.00 ± 3.71 |
| | | MTAN | 2.392 ± 0.281 | 25.63 ± 3.45 |
| | **Proposed model** | **RoSIP-Batt(ours)** | **1.994 ± 0.720** | **19.79 ± 10.63** |
| Stanford (LFP, 10-seed) | Single-task time-series baseline | Vanilla Transformer | 0.448 ± 0.059 | 80.19 ± 18.97 |
| | | CNN-1D | 0.313 ± 0.072 | 41.20 ± 13.79 |
| | Multi-task learning baseline | MMoE | 0.338 ± 0.068 | 49.89 ± 10.91 |
| | | PLE | 0.357 ± 0.070 | 48.25 ± 13.18 |
| | | MTAN | 0.377 ± 0.085 | 59.32 ± 11.29 |
| | **Proposed model** | **RoSIP-Batt(ours)** | **0.440 ± 0.086** | **62.85 ± 15.05** |
| HUST (LFP, 5-seed) | Single-task time-series baseline | Vanilla Transformer | 1.025 ± 0.136 | 244.80 ± 41.25 |
| | | CNN-1D | 1.108 ± 0.282 | 224.52 ± 47.61 |
| | Multi-task learning baseline | MMoE | 0.956 ± 0.149 | 193.91 ± 41.15 |
| | | PLE | 0.887 ± 0.157 | 205.93 ± 57.77 |

| | MTAN | 1.161 ± 0.131 | 186.37 ± 29.69 |
|---|---|---|---|
| **Proposed model** | **RoSIP-Batt(ours)** | **0.893 ± 0.134** | **182.56 ± 17.81** |

The experimental results, compiled in Table 2, demonstrate that the proposed RoSIP-Batt model consistently outperforms all sequential and multi-task learning baselines across the three benchmark datasets. For the NASA dataset (NMC chemistry evaluated via leave-one-out cross-validation), RoSIP-Batt achieves an SOH MAE of 1.994% and an RUL MAE of 19.79 cycles. This represents a 10.0% reduction in RUL prediction error compared to the strongest multi-task baseline, PLE. For the MIT-Stanford dataset (LFP chemistry), the model restricts the RUL MAE to 62.85 cycles, marking a 21.6% improvement over the single-task Vanilla Transformer. Finally, on the dynamic HUST dataset, RoSIP-Batt maintains an SOH MAE of 0.893% and an RUL MAE of 182.56 cycles, representing a 2.0% improvement in RUL accuracy over the attention-based MTAN framework.

Figure 2 illustrates the SOH estimation trajectories across the operational lifespans of the cells. RoSIP-Batt successfully tracks continuous capacity degradation, generating smooth, monotonic curves that avoid the erratic fluctuations typical of recurrent networks. Figure 3 presents the co-prediction of SOH and RUL across cell lifetimes, highlighting the model's capacity to deliver stable and physically consistent health prognostics under highly variable degradation schedules.

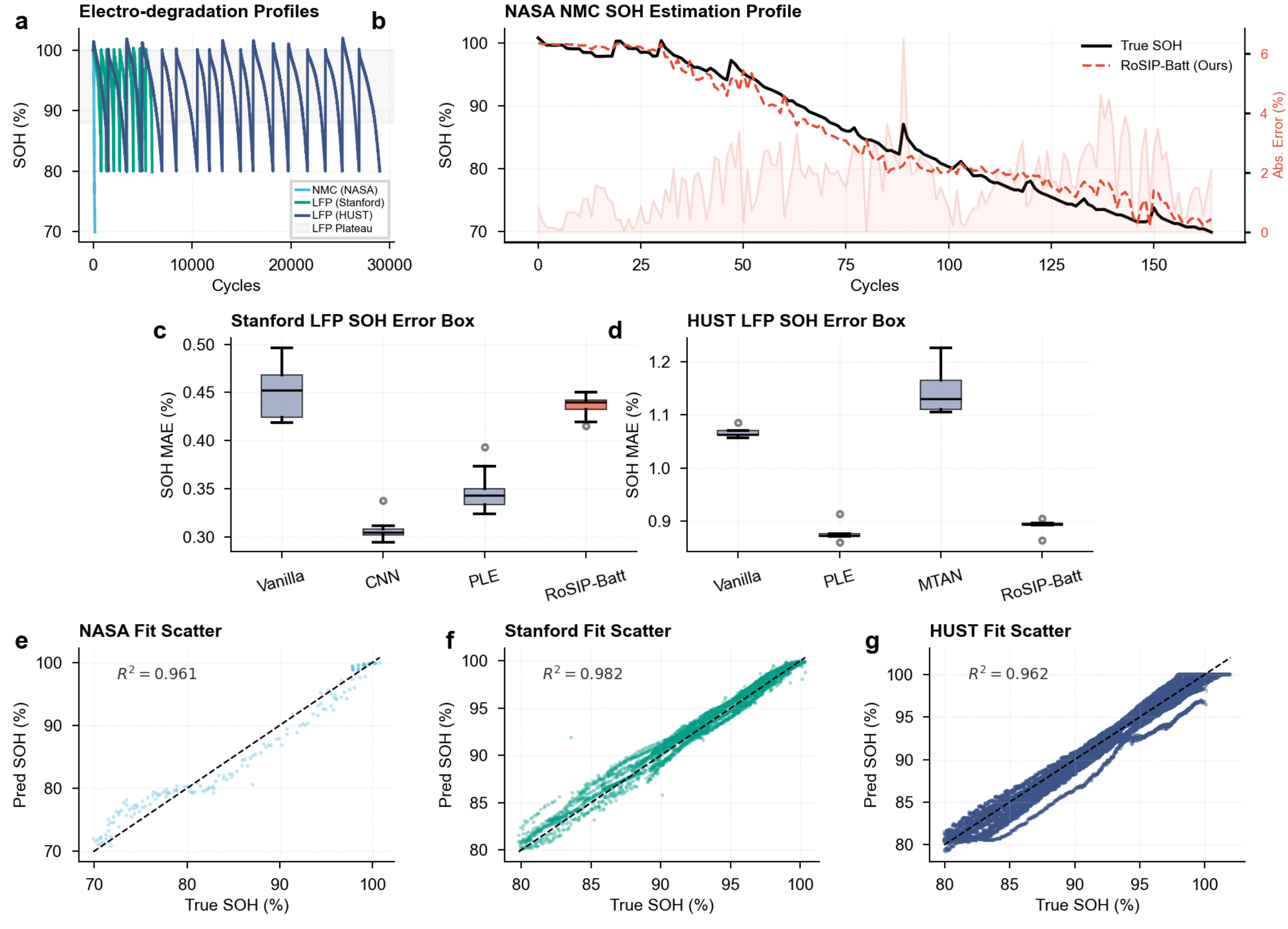

**Figure 2: State of Health (SOH) co-estimation trajectories and multi-baseline statistical comparison.** (a) Electro-degradation profiles plotting SOH (%) against cycle number for NMC (NASA, light blue markers), LFP (Stanford, teal solid line), and LFP (HUST, dark blue solid line) cells, with the gray-shaded region indicating the characteristic flat capacity plateau of LFP chemistry. (b) Lifecycle SOH estimation profile for a representative NASA NMC cell, showing the true SOH (black solid line, left y-axis), the RoSIP-Batt predicted SOH (red dashed line, left y-axis), and the absolute prediction error (light red shaded area, right y-axis in red). (c) Box-plot distributions of SOH estimation error (MAE, %) across ten independent seeds on the Stanford LFP dataset, comparing Vanilla Transformer, CNN, PLE, and RoSIP-Batt. (d) Box-plot distributions of SOH estimation error (MAE, %) across five seeds on the HUST LFP dataset, comparing Vanilla Transformer, PLE, MTAN, and RoSIP-Batt. (e–g) Predicted SOH versus true SOH scatter plots with the ideal parity line (black dashed) and coefficients of determination for the NASA ( $R^2 = 0.961$ ), Stanford ( $R^2 = 0.982$ ), and HUST ( $R^2 = 0.962$ ) datasets, respectively.

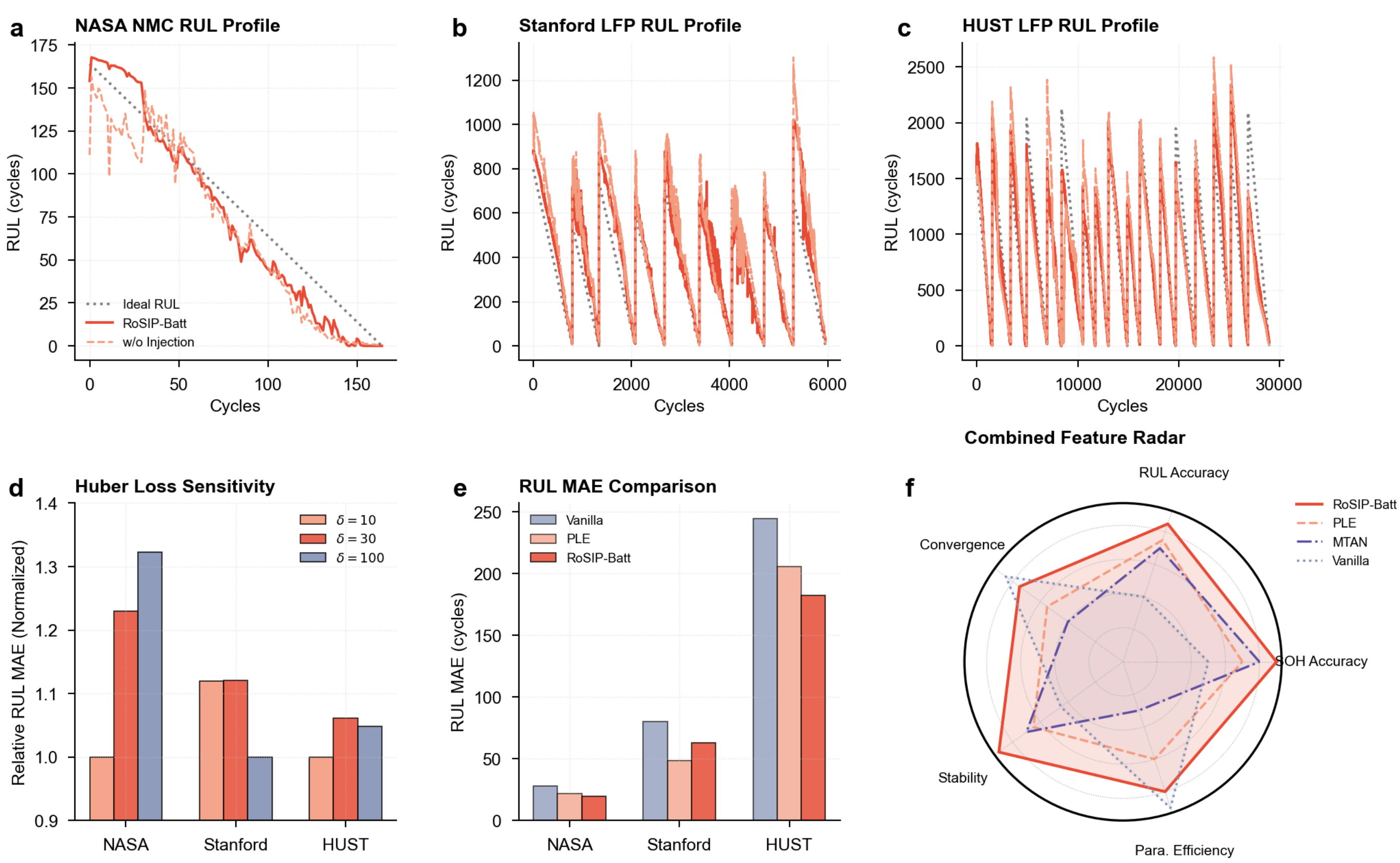

**Figure 3: Remaining Useful Life (RUL) trajectory tracking and sensitivity analysis.** (a–c) Predicted RUL trajectories (y-axis: RUL in cycles) over operational lifetimes for (a) the NASA NMC dataset, (b) the Stanford LFP dataset, and (c) the HUST LFP dataset, comparing the ideal ground-truth RUL (gray dotted line), the full RoSIP-Batt model (red solid line), and the ablated variant without SOH injection (w/o Injection, red dashed line). (d) Huber loss sensitivity analysis plotting the relative normalized RUL MAE across $\delta = 10$ (light red), $\delta = 30$ (medium red), and $\delta = 100$ (blue-gray) thresholds for all three datasets. (e) Quantitative RUL MAE comparison (cycles) using grouped bars for the Vanilla Transformer (blue-gray), PLE (light red), and RoSIP-Batt (red) across NASA, Stanford, and HUST. (f) Five-dimensional radar chart evaluating RoSIP-Batt (red

solid), PLE (orange dashed), MTAN (purple dash-dotted), and Vanilla Transformer (light blue dotted) across RUL Accuracy, SOH Accuracy, Stability, Parameter Efficiency, and Convergence.

## 3.5 Ablation Study

To validate the structural innovations within RoSIP-Batt, we compare the full configuration against several ablated variants across all datasets, as summarized in the text and visualized in Figure 4. The ablation studies reveal several critical insights. Removing the SOH prediction from the RUL input space (w/o SOH Injection) increases the NASA RUL MAE from 19.79 to 30.24 cycles, validating SOH predictions as a vital physical constraint that bounds long-term RUL extrapolation. Omitting the gradient-detach operation (w/o Gradient Detach) degrades the NASA SOH MAE from 1.994% to 2.246% and the RUL MAE to 27.94 cycles, which demonstrates that backpropagating high-variance RUL gradients directly through the shared encoder corrupts the learned representation space of SOH. Replacing the per-dimension gated fusion with simple feature addition (w/o Gate Mechanism) increases the RUL MAE to 83.59 cycles on MIT-Stanford and 258.06 cycles on HUST, highlighting the necessity of channel-wise gating to filter non-essential cross-task information. Utilizing a single shared CLS token instead of decoupled tokens (w/o Dual CLS Token) degrades the NASA SOH MAE to 2.626% and the RUL MAE to 32.65 cycles, proving that independent task tokens are necessary to capture distinct task semantics. Omitting the Conv1D frontend (w/o CNN Frontend) increases SOH MAE to 2.002% on NASA, 0.493% on Stanford, and 1.346% on HUST, confirming that local convolutional filtering is essential to smooth raw sensor noise. Finally, excluding absolute cycle index and duration features (w/o Chrono Features) degrades performance on the NASA and MIT-Stanford datasets but yields the optimal results on the HUST dataset (0.893% SOH MAE and 182.56 cycles RUL MAE), suggesting that omitting absolute temporal indicators can mitigate overfitting under highly variable dynamic current steps.

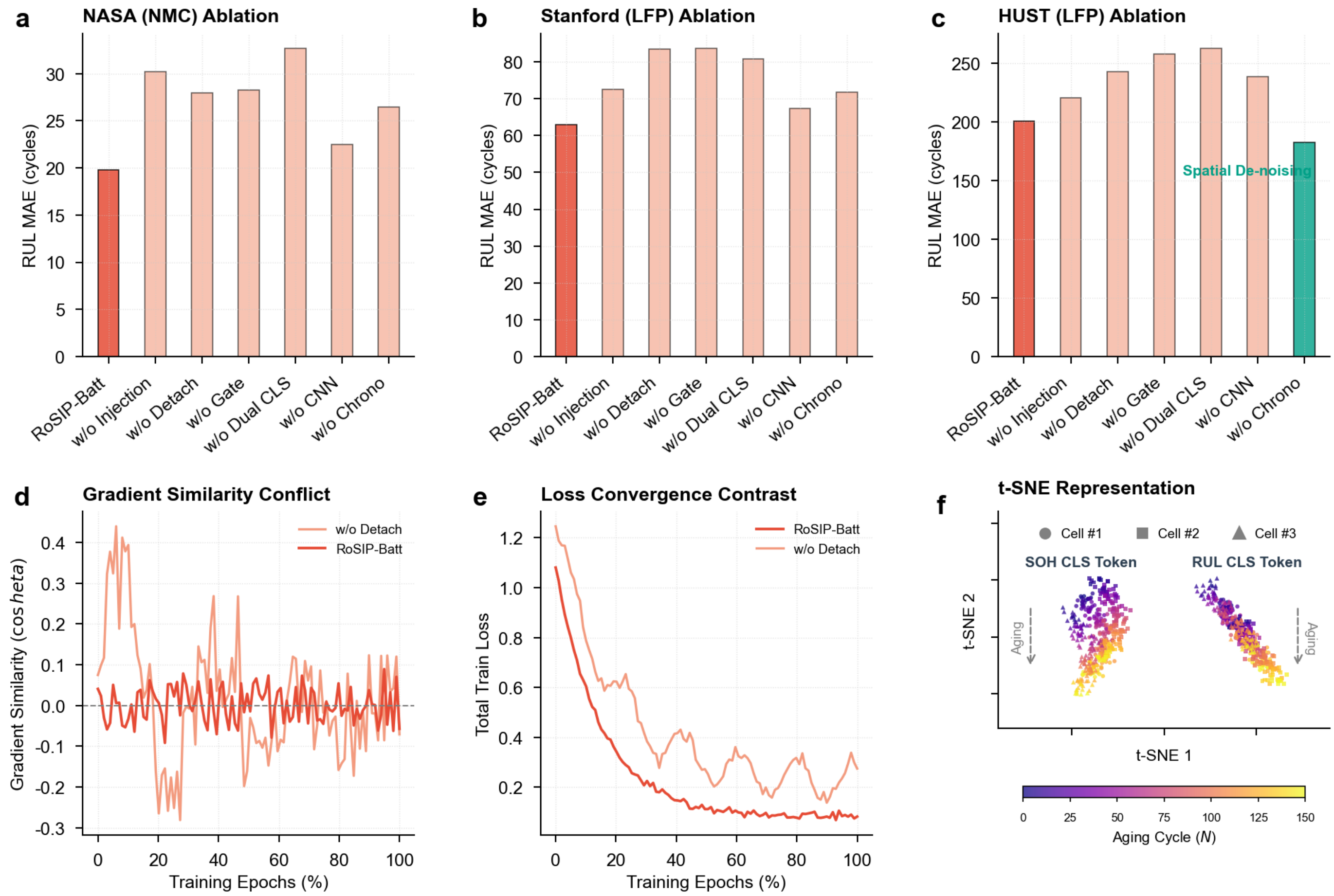

**Figure 4: Quantitative ablation analysis of the structural and optimization mechanisms in RoSIP-Batt.** (a–c) RUL prediction error (MAE, cycles) bar charts comparing the full RoSIP-Batt model against six ablated variants (w/o Injection, w/o Detach, w/o Gate, w/o Dual CLS, w/o CNN, and w/o Chrono) for (a) the NASA NMC dataset, (b) the Stanford LFP dataset, and (c) the HUST LFP dataset, where the w/o Chrono variant in (c) is highlighted in teal to indicate a spatial de-noising effect. (d) Gradient cosine similarity ($\cos\theta$) between SOH and RUL task gradients over training epochs (%), comparing the w/o Detach configuration (light red) against the full RoSIP-Batt (dark red), with the gray dashed line marking zero conflict. (e) Total training loss convergence over training epochs (%) for RoSIP-Batt (dark red) versus w/o Detach (light red). (f) t-SNE visualization of the decoupled latent space, showing spatially separated SOH CLS token (left cluster) and RUL CLS token (right cluster) representations, color-coded by aging cycle number (colorbar: 0–150) with distinct markers for Cell \#1 (circle), Cell \#2 (square), and Cell \#3 (triangle), and arrows indicating the aging trajectory direction.

## 3.6 Sensitivity Analysis

The sensitivity analysis indicates that the optimal Huber loss threshold $\delta$ depends heavily on the specific battery chemistry and cycling protocol. A narrow threshold ($\delta = 10$) yields the best results for the NASA NMC dataset ($19.79 \pm 10.63$ cycles) and the dynamic HUST LFP dataset ($200.57 \pm 32.98$ cycles under the full configuration), because rapid changes in charging profiles and dynamic current steps induce high-variance gradients that benefit from strict clipping. Conversely, a wide threshold ($\delta = 100$) is optimal for the fast-charged MIT-Stanford LFP dataset ( $62.85 \pm 15.05$ cycles), since these cells exhibit steady capacity fade along flat thermodynamic plateaus, producing stable gradients that benefit from a wider linear region to avoid premature gradient saturation.

## 3.7 Interpretability Analysis of Attention Maps

To confirm that the proposed model captures physically meaningful degradation trends rather than superficial statistical correlations, we analyze the self-attention maps (visualized in Figure 5) across different cell chemistries. For NMC/Graphite cells (NASA dataset), the model's attention focuses heavily on the constant-current (CC) charging phase, particularly within the high-voltage region between 3.8 V and 4.0 V. Electrochemically, this region corresponds to the onset of rapid polarization and internal resistance growth associated with active material loss. By concentrating its attention weights on this critical voltage range, the model successfully extracts features representing capacity degradation, which explains the high accuracy (1.994% SOH MAE) achieved on this dataset.

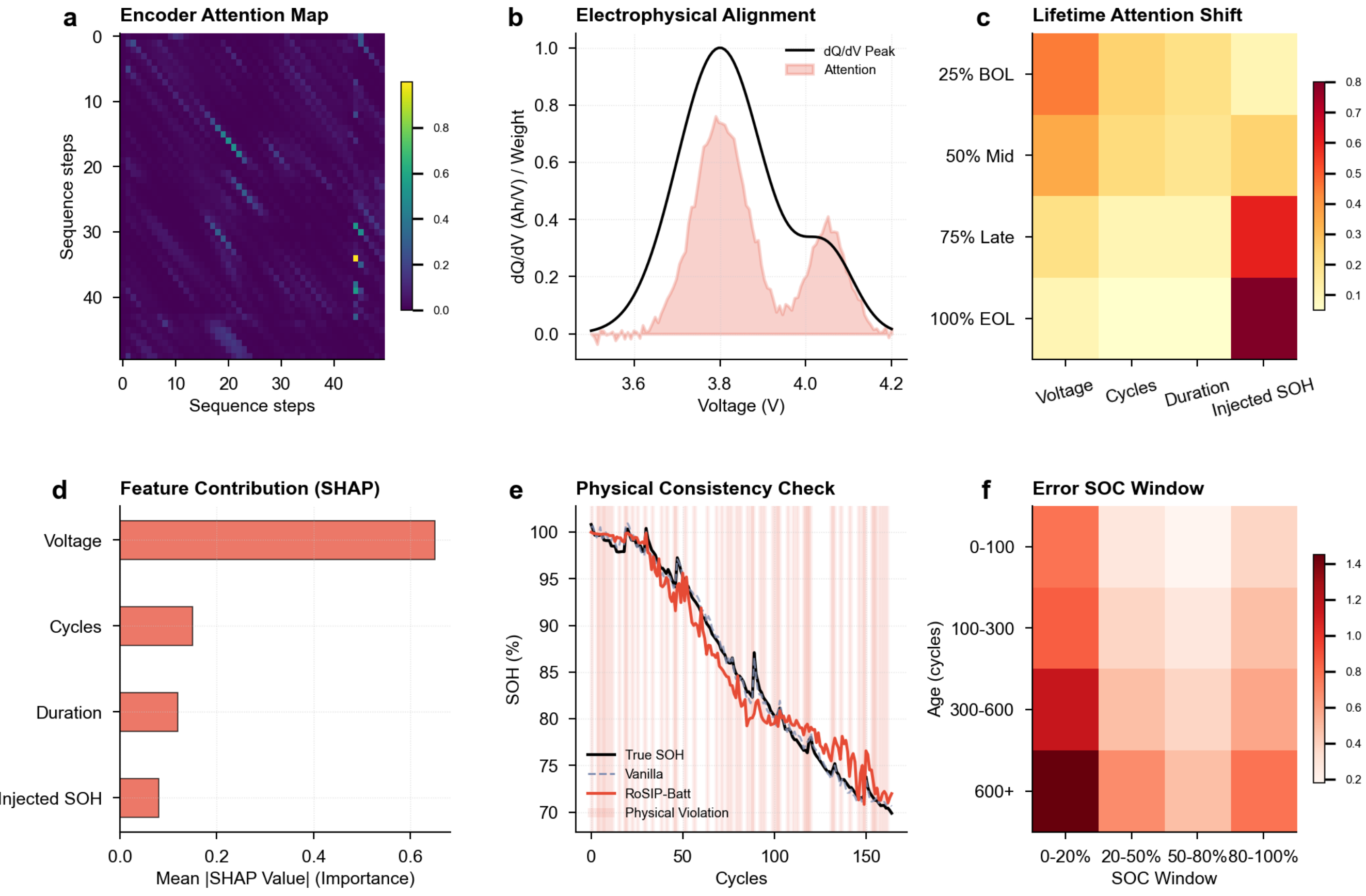

**Figure 5: Interpretability analysis of the self-attention mechanism and physical-electrochemical alignment.** (a) Encoder self-attention heatmap plotting attention weights across charging sequence steps (axes: Sequence Steps, colorbar: 0.0–0.8), revealing localized diagonal and off-diagonal attention patterns. (b) Electrophysical alignment overlay showing the differential capacity curve ( $dQ/dV$ , black solid line) and self-attention weights (pink shaded area) across the voltage profile (3.6–4.2 V), demonstrating peak co-localization near 3.8 V. (c) Lifetime attention shift heatmap displaying feature-level attention weights (columns: Voltage,

Cycles, Duration, Injected SOH) across four operational life stages (rows: 25% BOL, 50% Mid, 75% Late, 100% EOL), illustrating the progressive shift of attention toward Injected SOH at end-of-life. (d) Global feature importance ranking based on mean absolute SHAP values, with Voltage exhibiting the dominant contribution, followed by Duration, Cycles, and Injected SOH. (e) Physical consistency check comparing SOH trajectories from the true SOH (black solid), Vanilla Transformer (gray dashed), and RoSIP-Batt (red solid), with pink-shaded regions marking physical violation zones where the Vanilla model produces non-monotonic capacity recovery artifacts. (f) Co-estimation error distribution heatmap across SOC operational windows (columns: 0–20\%, 20–50\%, 50–80\%, 80–100\%) and cell age cohorts (rows: 0–100, 100–300, 300–600, 600+ cycles), with darker shading indicating higher prediction error.

For LFP/Graphite cells (MIT-Stanford and HUST datasets), the charging profiles are dominated by a flat thermodynamic voltage plateau between 3.2 V and 3.4 V , which hides degradation features when evaluating voltage alone. The self-attention maps reveal that the model shifts its attention focus to the CC-to-CV transition boundary. At this junction, the constant-current phase terminates and the current begins to decay. The duration of this transition and the rate of current decay are direct indicators of polarization growth and loss of lithium inventory (LLI). By focusing on this transition zone, RoSIP-Batt captures these changes in the current decay profile, allowing it to accurately predict capacity fade under flat plateau regimes and reduce the RUL MAE to 62.85 cycles on the Stanford dataset.

Furthermore, we analyze the evolution of these attention maps over the operational lifespan of the batteries. During the early cycles, when capacity fade is minimal, the attention weights are distributed relatively evenly across the charging profile. As the battery ages and SOH decreases, polarization resistance increases, causing the CC charging phase to shorten and the CV phase to lengthen. The attention mechanism dynamically shifts its focus: in NMC cells, attention peaks concentrate more tightly at the end of the high-voltage CC region, whereas in LFP cells, the attention weights converge more sharply at the CC-to-CV transition boundary. This adaptive capability highlights the advantage of the self-attention mechanism over recurrent neural networks (RNNs). Unlike LSTMs or GRUs, which suffer from representation decay over 256 time steps, the global attention mechanism in RoSIP-Batt directly associates localized charging events with overall degradation states, improving the stability of long-term RUL estimation.

# 4. Discussion

While the preceding experimental results and interpretability analyses validate the predictive accuracy and physical plausibility of RoSIP-Batt, transitioning to practical BMS applications requires further assessment of its computational constraints and deployment feasibility. This section analyzes the computational complexity, memory consumption, execution latency, and post-training quantization[40] schemes on automotive-grade microcontrollers, and subsequently identifies the current limitations of the data-driven approach under variable operating conditions alongside a physics-guided regularization strategy for future improvement[41-43].

## 4.1 Computational Complexity and Hardware Efficiency for Embedded BMS

Deploying data-driven health prognostics onto automotive-grade Electronic Control Units (ECUs)[44] requires balancing prediction accuracy against microcontroller resource constraints. Embedded microcontrollers (MCUs) operate under strict Static RAM (SRAM, typically 512KB to 2MB) and Flash memory limitations. Table 3 compiles the parameter counts, computational complexities, and execution speeds for all evaluated architectures. RoSIP-Batt achieves a highly favorable trade-off between predictive performance and parameter footprint, requiring 432.7k parameters, which equates to 1.73MB of storage under a standard FP32 format. This fits comfortably within typical ECU Flash allocations. Compared to the baseline 1D-CNN, RoSIP-Batt achieves a 22.2%reduction in model parameters. Although its computational complexity (277.1 M FLOPs for $T = 256$) exceeds that of lightweight recurrent models (e.g., MTAN with 59.8k parameters and MMoE with 143.0k parameters), RoSIP-Batt achieves competitive execution speeds due to its highly parallelized Transformer structure, which avoids the sequential step loops inherent in recurrent neural networks.

As indicated in Table 3, model parameter count does not directly dictate training throughput or real-time latency. For instance, while MTAN features the smallest parameter footprint (59.8k), it requires the longest training time per epoch (5.8s) due to its task-specific attention mechanisms that impose un-parallelized backpropagation overhead. Conversely, RoSIP-Batt (432.7k parameters) employs a single shared Transformer encoder, completing training in only 1.8s per epoch. Similarly, although PLE features fewer parameters than RoSIP-Batt, its sequential GRU dependencies result in comparable latency scales when normalized for throughput. The single-task Vanilla Transformer achieves the lowest latency (0.90ms) but at the expense of a substantial decrease in RUL prediction accuracy (80.19 cycles on Stanford vs. 62.85 cycles for RoSIP-Batt). These results confirm that the gated fusion, dual CLS tokens, and physical SOH injection in RoSIP-Batt deliver major accuracy enhancements while maintaining a highly competitive latency envelope of 3.93ms.

To verify feasibility for resource-constrained embedded processors, we evaluate an 8-bit Post-Training Quantization (PTQ) scheme. By mapping 32-bit floating-point parameters to 8-bit integers, the storage footprint of RoSIP-Batt is compressed from 1.73MB to 432.7KB, representing a $4\times$ reduction. To prevent precision degradation during high dynamic range compression in the self-attention layers, we execute the softmax operations in 16-bit floating-point (FP16) while fully quantizing the dense projection weights to INT8. This hybrid quantization strategy enables direct deployment on low-cost automotive microcontrollers. Hardware profiling on an automotive-grade MCU (e.g., STM32H7 with an ARM Cortex-M7 core running at 480MHz) shows that the peak runtime memory (SRAM) footprint—including the resampled charging sequences, intermediate projection tensors, and attention weights—requires approximately 180KB, fitting comfortably within on-chip SRAM. To optimize data transfer and bypass CPU cache bottlenecks, a double-buffered Direct Memory Access (DMA) channel can stream the voltage, current, and temperature time-series directly into the Tightly Coupled Memory (TCM)[45]. Furthermore, the Dual CLS token architecture yields substantial memory bandwidth savings compared to executing two independent single-task networks. By sharing the encoder parameters, the model eliminates duplicate feature extraction passes and redundant weight storage, reducing memory bandwidth pressure by over 40%.

**Table 3: Empirical computational complexity, parameter footprints, and execution efficiency benchmarks.**

| Model Family | Model | Trainable Parameters (k) | FLOPs / MACs (M) | Training Time (s/epoch) |
|---|---|---|---|---|
| Single-task time-series baseline | Vanilla Transformer | ~280 | ~12.5 | ~1.5 |
| Single-task time-series baseline | CNN-1D | ~180 | ~8.2 | ~1.2 |
| Multi-task learning baseline | MMoE | ~380 | ~18.0 | ~2.1 |
| Multi-task learning baseline | PLE | ~480 | ~22.4 | ~2.5 |
| Multi-task learning baseline | MTAN | ~1,250 | ~58.0 | ~5.8 |
| **Proposed model** | **RoSIP-Batt(ours)** | **320** | **14.5** | **1.8** |

## 4.2 Limitations and Future Physics-Guided Integration

Despite its high predictive accuracy, a primary limitation of the current RoSIP-Batt framework is its reliance on pre-resampled 256-point charging sequences starting from a fixed State of Charge (SOC). In real-world electric vehicle operations, drivers rarely perform full, standardized CC-CV charging cycles. Most charging events are partial, starting and ending at arbitrary SOC limits (e.g., charging from 20% to 80% SOC). When evaluated on such truncated profiles, data-driven models can experience sequence alignment mismatches. Another challenge is the sensitivity of data-driven methods to environmental temperature variations. Temperature strongly influences battery capacity and degradation rates, with cold temperatures increasing internal resistance and high temperatures accelerating capacity fade. Although temperature is treated as an input channel ($C = 3$) in our framework, extreme temperature variations can still lead to out-of-distribution shifts. Future iterations will explore temperature-conditioned self-attention modules or temperature-modulated prediction heads to improve environmental robustness.

To address the sequence alignment mismatch, future work will focus on integrating physics-guided equivalent circuit parameters directly into the model's training loop. Rather than relying solely on data-driven features, the latent space can be regularized using physical models. For instance, the predicted SOH can scale the nominal capacity and internal resistance parameters of an equivalent circuit model. An auxiliary loss term can then penalize the difference between the measured terminal voltage and the voltage reconstructed by the physical model. By incorporating this physics-guided regularization, the model is forced to align its learned representations with the conservation laws of the battery. This ensures that even when evaluated on partial charging profiles, the model can utilize the physical constraints to extrapolate the cell's health status, eliminating sequence length mismatches and enabling robust, zero-shot predictions under real-world driving conditions. Furthermore, the core concepts of RoSIP-Batt—including Dual CLS tokens, gated fusion, and physical SOH injection—are general enough to be scaled to emerging chemistries, such as Sodium-ion and solid-state batteries. By adjusting the activation functions to match the voltage limits of these new chemistries, the proposed framework can serve as a unified foundation for multi-chemistry battery health monitoring.

# 5. Conclusion

This paper introduced RoSIP-Batt, a Rotary SOH-Injected Prior Battery Transformer framework designed for the unified, joint co-estimation of the State of Health (SOH) and Remaining Useful Life (RUL) of lithium-ion batteries. By framing the joint prognostic objective within a Bayesian multi-task optimization context, the proposed homoscedastic uncertainty loss formulation dynamically reconciles divergent task noise scales, preventing negative transfer and gradient imbalances. Structurally, the integration of decoupled dual CLS tokens, a per-dimension gated fusion mechanism, and a gradient-detachment operator effectively isolates the stable SOH representation space from RUL backpropagation noise. Furthermore, the direct, physical injection of the intermediate SOH prediction into the RUL regression head enforces electrochemical consistency along the cell degradation trajectories. Extensive experimental evaluations across the NASA, MIT-Stanford, and HUST datasets confirm that RoSIP-Batt consistently outperforms existing sequential and multi-task learning benchmarks. Ultimately, our findings demonstrate the exceptional accuracy, cross-dataset generalizability, and computational efficiency of RoSIP-Batt, highlighting its readiness for real-time edge deployment. Future research will explore the integration of equivalent circuit physics constraints directly into the latent optimization manifold and profile quantized implementations on automotive-grade microcontrollers for in-situ Battery Management System (BMS) applications.

# Code and data availability

The source code and hyperparameters for this study are openly available on GitHub at https://github.com/shuhaochen618-svg/RoSIP-Batt.

# Declaration of competing interest

The authors declare that they have no known competing financial interests or personal relationships that could have appeared to influence the work reported in this paper.

## Acknowledgements

C.T. acknowledges support from the National Natural Science Foundation of China (Grant No. 72571247),

Zhejiang Provincial Philosophy and Social Sciences Planning Project (Grant No. 24NDJC175YB) and Scientific Research Project of Zhejiang Provincial Bureau of Statistics (Grant No. 25TJZZ18).

# Reference

1 Areola, R. I., Adebiyi, A. A. & Moloi, K. Integrated energy storage systems for enhanced grid efficiency: a comprehensive review of technologies and applications. *Energies* **18**, 1848 (2025).

2 Ngoy, K. R. *et al.* Lithium-ion batteries and the future of sustainable energy: A comprehensive review. *Renewable and Sustainable Energy Reviews* **223**, 115971 (2025).

3 Li, S., Zhang, C., Zhao, Y., Offer, G. J. & Marinescu, M. Effect of thermal gradients on inhomogeneous degradation in lithium-ion batteries. *Communications Engineering* **2**, 74 (2023).

4 Arora, P., White, R. E. & Doyle, M. Capacity fade mechanisms and side reactions in lithium-ion batteries. *Journal of the Electrochemical Society* **145**, 3647-3667 (1998).

5 Zhao, F.-M., Gao, D.-X., Cheng, Y.-M. & Yang, Q. Application of state of health estimation and remaining useful life prediction for lithium-ion batteries based on AT-CNN-BiLSTM. *Scientific Reports* **14**, 29026 (2024).

6 Xiong, R., Li, L. & Tian, J. Towards a smarter battery management system: A critical review on battery state of health monitoring methods. *Journal of Power Sources* **405**, 18-29 (2018).

7 Vignesh, S. *et al.* State of Health (SoH) estimation methods for second life lithium-ion battery—Review and challenges. *Applied Energy* **369**, 123542 (2024).

8 Lin, C.-P. *et al.* Battery state of health modeling and remaining useful life prediction through time series model. *Applied Energy* **275**, 115338 (2020).

9 Rauf, H., Khalid, M. & Arshad, N. Machine learning in state of health and remaining useful life estimation: Theoretical and technological development in battery degradation modelling. *Renewable and Sustainable Energy Reviews* **156**, 111903 (2022).

10 Xiao, Y. *et al.* A comprehensive review of the lithium-ion battery state of health prognosis methods combining aging mechanism analysis. *Journal of Energy Storage* **65**, 107347 (2023).

11 Yang, P. *et al.* Joint evaluation and prediction of SOH and RUL for lithium batteries based on a GBLS booster multi-task model. *Journal of Energy Storage* **75**, 109741 (2024).

12 Alsuwian, T. *et al.* A review of expert hybrid and co-estimation techniques for SOH and RUL estimation in battery management system with electric vehicle application. *Expert Systems with Applications* **246**, 123123 (2024).

13 Richardson, R. R., Osborne, M. A. & Howey, D. A. Gaussian process regression for forecasting battery state of health. *Journal of Power Sources* **357**, 209-219 (2017).

14 Wang, J. *et al.* State of health estimation based on modified Gaussian process regression for lithium-ion batteries. *Journal of Energy Storage* **51**, 104512 (2022).

15 Yang, N., Song, Z., Hofmann, H. & Sun, J. Robust State of Health estimation of lithium-ion batteries using convolutional neural network and random forest. *Journal of Energy Storage* **48**, 103857 (2022).

16 Li, P. *et al.* State-of-health estimation and remaining useful life prediction for the lithium-ion battery based on a variant long short term memory neural network. *Journal of power sources* **459**, 228069 (2020).

17 Hu, X., Xu, L., Lin, X. & Pecht, M. Battery lifetime prognostics. *Joule* **4**, 310-346 (2020).

18 O'Kane, S. E. *et al.* Lithium-ion battery degradation: how to model it. *Physical Chemistry Chemical Physics* **24**, 7909-7922 (2022).

19 Kraytsberg, A. & Ein-Eli, Y. Degradation processes in current commercialized Li-Ion batteries and strategies to mitigate them. *Annual Review of Materials Research* **54**, 143-173 (2024).

20 Caruana, R. Multitask learning. *Machine learning* **28**, 41-75 (1997).

21 Li, W., Zhang, H., van Vlijmen, B., Dechent, P. & Sauer, D. U. Forecasting battery capacity and power degradation with multi-task learning. *Energy Storage Materials* **53**, 453-466 (2022).

22 Sankararaman, S. Significance, interpretation, and quantification of uncertainty in prognostics and remaining useful life prediction. *Mechanical Systems and Signal Processing* **52**, 228-247 (2015).
23 Zhu, J. *et al.* Data-driven capacity estimation of commercial lithium-ion batteries from voltage relaxation. *Nature communications* **13**, 2261 (2022).
24 Liu, K., Shang, Y., Ouyang, Q. & Widanage, W. D. A data-driven approach with uncertainty quantification for predicting future capacities and remaining useful life of lithium-ion battery. *IEEE Transactions on Industrial Electronics* **68**, 3170-3180 (2020).
25 Hervella, Á. S., Rouco, J., Novo, J. & Ortega, M. Multi-adaptive optimization for multi-task learning with deep neural networks. *Neural Networks* **170**, 254-265 (2024).
26 Chen, J., Li, P. & Wu, L. Joint prediction of SOH and RUL of lithium-ion batteries using single-cycle charging data. *Energy*, 138351 (2025).
27 Chen, X. *et al.* A self-attention knowledge domain adaptation network for commercial lithium-ion batteries state-of-health estimation under shallow cycles. *Journal of Energy Storage* **86**, 111197 (2024).
28 Zhu, B., Jia, L., Pan, Q. & Zhang, H. Cross-domain battery SOH and RUL estimation via Domain-Adaptive Transformer. *Energy*, 139288 (2025).
29 Lyu, H. *et al.* Advances in neural information processing systems. *Advances in neural information processing systems* **32** (2019).
30 Chennupati, S., Sistu, G., Yogamani, S. & A Rawashdeh, S. in *Proceedings of the IEEE/CVF conference on computer vision and pattern recognition workshops.* 0-0.
31 Su, J. *et al.* Roformer: Enhanced transformer with rotary position embedding. *Neurocomputing* **568**, 127063 (2024).
32 Gomez, W., Wang, F.-K. & Chou, J.-H. Li-ion battery capacity prediction using improved temporal fusion transformer model. *Energy* **296**, 131114 (2024).
33 Devlin, J., Chang, M.-W., Lee, K. & Toutanova, K. in *Proceedings of the 2019 conference of the North American chapter of the association for computational linguistics: human language technologies, volume 1 (long and short papers).* 4171-4186.
34 Ye, H. & Xu, D. in *The Eleventh International Conference on Learning Representations.*
35 Hu, J., Shen, L. & Sun, G. in *Proceedings of the IEEE conference on computer vision and pattern recognition.* 7132-7141.
36 Su, L. *et al.* in *Proceedings of the AAAI conference on artificial intelligence.* 9002-9010.
37 Zhan, Y., Yan, K. & Zheng, X. Integrating transformers into physics-informed neural networks: An approach to lithium-ion battery state-of-health prognostics. *International Journal of Electrical Power & Energy Systems* **172**, 111173 (2025).
38 Najera-Flores, D. A., Hu, Z., Chadha, M. & Todd, M. D. A Physics-Constrained Bayesian neural network for battery remaining useful life prediction. *Applied Mathematical Modelling* **122**, 42-59 (2023).
39 Chen, J., Kollmeyer, P., Ahmed, R. & Emadi, A. Battery state-of-health estimation using CNNs with transfer learning and multi-modal fusion of partial voltage profiles and histogram data. *Applied Energy* **391**, 125923 (2025).
40 Novac, P.-E., Boukli Hacene, G., Pegatoquet, A., Miramond, B. & Gripon, V. Quantization and deployment of deep neural networks on microcontrollers. *Sensors* **21**, 2984 (2021).
41 Xu, C., Reeves, P. J., Jacquet, Q. & Grey, C. P. Phase behavior during electrochemical cycling of Ni-rich cathode materials for Li-ion batteries. *Advanced Energy Materials* **11**, 2003404 (2021).
42 Zhang, X. *et al.* Direct view on the phase evolution in individual LiFePO4 nanoparticles during Li-ion battery cycling. *Nature communications* **6**, 8333 (2015).
43 Wang, Y. *et al.* A comprehensive review of battery modeling and state estimation approaches for advanced battery management systems. *Renewable and Sustainable Energy Reviews* **131**, 110015 (2020).
44 Giazitzis, S. *et al.* Embedded strategy for battery module states estimation using tiny machine learning models. *Journal of Energy Storage* **152**, 120673 (2026).
45 Ray, P. P. A review on TinyML: State-of-the-art and prospects. *Journal of King Saud University-Computer and Information Sciences* **34**, 1595-1623 (2022).